\documentclass[times, review, 10pt]{elsarticle}

\usepackage[T1]{fontenc}
\usepackage{amssymb}
\usepackage{amsmath}
\usepackage{amsthm}
\usepackage{graphicx}
\usepackage{booktabs}
\usepackage{multirow}
\usepackage[table,xcdraw]{xcolor}
\usepackage{colortbl}
\usepackage{float}
\usepackage{siunitx}
\usepackage{pifont}
\usepackage{url}
\usepackage{textcomp}
\usepackage{subfig}
\usepackage{algorithm}
\usepackage{algorithmic}
\usepackage[hidelinks]{hyperref}
\usepackage{enumitem}
\usepackage[font=small,labelfont=bf,skip=6pt]{caption}
\usepackage{bbm}
\usepackage{placeins}

\definecolor{MyBlue}{HTML}{00B0F0}
\definecolor{MyRed}{HTML}{FF0000}

\newcommand{\cmark}{\ding{51}}
\newcommand{\xmark}{\ding{55}}

\journal{Pattern Recognition}

\begin{document}

\begin{frontmatter}

\title{UniFLM: United Segmentation and Measurement on Fetal Limb Ultrasonic Image}

\author[1]{Zeen Zhou\fnref{fn1}}
\ead{chosen1203@whu.edu.cn}

\author[2]{Qiuhua Chen\fnref{fn1}}
\ead{qiuhuachen@whu.edu.cn}

\author[4]{Xiaojun Cao}
\author[4]{Changmao Chen}

\author[2,3]{\newline Chao Sun\corref{cor1}}
\ead{chaosun@whu.edu.cn}

\author[2,3]{Bo Du\corref{cor1}}
\ead{dubo@whu.edu.cn}

\address[1]{Academy of Advanced Interdisciplinary Studies, Wuhan University, Wuhan, China}
\address[2]{School of Computer Science, Wuhan University, Wuhan, China}
\address[3]{School of Computer Science, National Engineering Research Center for Multimedia Software, and Hubei Key Laboratory of Multimedia and Network Communication Engineering, Institute of Artificial Intelligence, Wuhan University, Wuhan 430072, China}
\address[4]{Guangzhou Women and Children's Medical Center, Guangzhou, China}

\fntext[fn1]{These authors contributed equally to this work.}
\cortext[cor1]{Corresponding authors.}

\begin{abstract}
Prenatal ultrasound examination is crucial for assessing fetal limb development and detecting congenital anomalies. However, existing artificial intelligence models often overlook fetal lethal skeletal dysplasias due to the lack of high-quality annotated data and a unified framework for multiple long bones. Moreover, generic segmentation models struggle with the inherent noise and semantic gaps in ultrasound images. To address these challenges, we construct the Fetal Limb Bones (FLB) dataset, comprising high-quality annotations for the humerus, femur, tibia-fibula, and radius-ulna. Furthermore, we propose UniFLM (United Segmentation and Measurement on Fetal Limb Ultrasonic Image), a unified framework for automatic cross-plane segmentation and measurement. UniFLM incorporates a Semantic Alignment Skip Connection (SASC) module to bridge the semantic gap between encoder and decoder features, and a Positive Sampling (PoSamp) strategy to filter noise and extract essential semantic information. Finally, a Point Regression Mapping (PRM) module is introduced to learn clinician annotation patterns for precise bone length measurement. Extensive experiments conducted on the FLB dataset and the public FetalP5 benchmark demonstrate that UniFLM achieves competitive performance with consistent generalization across four bone categories and external multi-center data, supported by comprehensive statistical validation including Bland-Altman agreement analysis and bootstrap confidence intervals. The source code is publicly available at \textcolor{blue}{\url{https://github.com/chosen1203/UniFLM}}.
\end{abstract}


\end{frontmatter}



\section{Introduction}
Prenatal ultrasound imaging is a crucial tool for assessing fetal anatomical structures and monitoring growth and development. Assessing the morphology and length of the fetal limb long bones, including the humerus, femur, leg (tibia-fibula), and forearm (radius-ulna), is clinically significant for diagnosing lethal skeletal dysplasias associated with severe limb shortening \cite{dighe2008fetal,salomon2011practice}. According to ISUOG guidelines and clinical practice \cite{salomon2019isuog,salomon2022isuog}, lethal skeletal dysplasias result in a poor postnatal prognosis, underscoring the importance of early prenatal diagnosis \cite{Nishimura2023Prenatal}. Compared to traditional diagnostic methods based on clinician expertise, artificial intelligence offers advantages of accuracy, speed, and automation, providing novel technological solutions for the intelligent diagnosis of severe congenital anomalies during the critical stages of pregnancy.

However, existing medical diagnosis models and datasets lack sufficient focus on limb long bone measurements and intelligent early diagnosis of fetal skeletal dysplasia, and face major challenges in clinical practice \cite{Carlson2017,krakow2009guidelines}. Specifically, as highlighted in early screening studies \cite{Schramm2009,Tretter1998}, there are three primary obstacles:

(1) Challenges posed by low-resolution, uneven contrast, and inherent noise in ultrasound images, as well as the coexistence of multiple bones in a single frame. These factors make it difficult to establish a robust and unified segmentation framework for fetal limb long bones in routine clinical settings.

(2) Lack of high-quality annotated data for fetal limb bones. The FPUS23 dataset \cite{Prabakaran2023FPUS23} only provides bounding box annotations, precluding precise measurement. The DeepGA model \cite{Dan2023DeepGA} predicts gestational age from femur length but neglects other long bones. Furthermore, \textbf{no public dataset exists with keypoint-level annotations for fetal limb bones}, a well-documented gap \cite{discover2025review}.

(3) Limitations of existing models in medical image segmentation tasks. The convolutional nature of U-Net-based \cite{Ronneberger2015Unet,siddique2021u} restricts its ability to capture long-range dependencies between features. Furthermore, the skip-connection mechanism is directly used to fuse the features between encoder and decoder blocks, which may introduce semantic gaps \cite{Ates2023Dual}. General-purpose foundation models like the Segment Anything Model (SAM) \cite{Kirillov2023Segment,mazurowski2023segment} and its medical adaptations \cite{zhang2023customized,zhang2024segment} still require extensive fine-tuning with high-quality data to achieve satisfactory performance.

To address the shortage of annotated ultrasound data for fetal skeletal analysis, we introduce the Fetal Limb Bones (\textbf{FLB}) dataset, encompassing four bone categories labeled by three senior clinicians. Building upon this benchmark, we propose an united segmentation and measurement model on fetal limb ultrasonic images (\textbf{UniFLM}) for end-to-end automated segmentation and biometric measurement of fetal limb long bones. The architecture integrates a U2-Net-based backbone with a Semantic Alignment Skip Connection (SASC) module and a Positive Sampling (PoSamp) mechanism. SASC bridges semantic gaps via dual cross-attention, while PoSamp suppresses acoustic noise. A Point Regression Mapping (PRM) strategy captures clinician-specific annotation patterns for precise measurement.

The primary contributions of this work are summarized as follows:
\begin{itemize}
\item We curate the \textit{Fetal Limb Bones (FLB)} dataset, a high-quality benchmark comprising multi-category ultrasound images annotated by senior experts.
\item We introduce the SASC module, which leverages dual cross-attention to explicitly align encoder and decoder features, bridging semantic discrepancies in ultrasound textures.
\item We develop a Positive Sampling (PoSamp) mechanism to suppress inherent ultrasound noise for robust feature extraction, coupled with a Point Regression Mapping (PRM) strategy for precise anatomical landmark localization.
\item Extensive evaluations, including \textbf{external validation on the public FetalP5 benchmark} (809 femur images from multiple centers), demonstrate UniFLM's generalization capability, supported by \textbf{comprehensive statistical analysis} (Bland-Altman agreement, bootstrap confidence intervals, Cohen's $d$ effect size). The FLB dataset and source code are publicly available at \textcolor{blue}{\url{https://github.com/chosen1203/UniFLM}}.
\end{itemize}


\section{Related Work}
\label{sec:related_work}

\subsection{Deep Learning in Medical Image Segmentation}
Deep learning has transformed medical image segmentation, driven by U-Net \cite{Ronneberger2015Unet} and variants (Attention U-Net \cite{Oktay2018}, UNet++ \cite{unetplusplus}, UNet 3+ \cite{huang2020unet}) with attention and dense skip connections. Vision Transformers (TransUNet \cite{transunet}, Swin-Unet \cite{swinunet}) capture global context but are computationally demanding. Foundation models (SAM \cite{Kirillov2023Segment}, MedSAM \cite{Ma2024MedSAM}, SAM-US \cite{lin2023samus}) offer generalization, while state space models (VM-UNet \cite{Ruan2024VMUNet}) and KANs (U-KAN \cite{Li2024UKAN}) provide efficient long-range modeling. Recent Pattern Recognition studies explore attention-based noise mitigation \cite{PR2023_attention, PR2024_medical}. Beyond ultrasound, CNN and U-Net-based frameworks have been successfully adapted to other modalities such as automated brain MRI segmentation \cite{verma2024mri}. Furthermore, recent surveys highlight that data scarcity remains a fundamental challenge in medical imaging \cite{discover2025review}, where limited annotated data constrains model performance---a challenge directly addressed by our FLB dataset. Despite these advances, generic architectures still struggle with severe acoustic noise in fetal ultrasound, motivating our SASC and PoSamp modules.

\subsection{Fetal Ultrasound Image Analysis}
Fetal ultrasound analysis primarily focuses on standard plane detection \cite{chen2015standard} and biometric measurement. While automated measurement of head (BPD, HC) and abdominal (AC) metrics is well-established \cite{avisdris2022biometrynet}, fetal limb assessment remains under-explored. Current resources for limb analysis are limited: the FPUS23 dataset \cite{Prabakaran2023FPUS23} provides only bounding box annotations, and models like DeepGA \cite{Dan2023DeepGA} focus exclusively on femur length, neglecting other long bones. A recent multi-center benchmark \cite{dromey2025multicentre} covers femur landmarks but was published after our study commenced. A unified framework that simultaneously segments and measures multiple limb bones remains a significant research gap.

\subsection{Landmark Detection and Biometric Measurement}
Biometric measurement relies on precise landmark localization, generally categorized into heatmap-based and regression-based methods. Heatmap approaches \cite{payer2019integrating} offer spatial uncertainty estimation but require computationally expensive post-processing. Direct coordinate regression methods \cite{toshev2014deeppose} are efficient but may lack spatial context. Building on this, our PRM module adopts a regression-based strategy enhanced by segmentation cues, learning to predict bone measurement endpoints by emulating clinician-specific annotation patterns. This ensures robust measurements even when anatomical boundaries are partially obscured by acoustic shadowing.


\section{Methodology}
\label{sec:method}

\subsection{Overview}
\label{sec:overview}

\begin{figure}[H]
\centering
\includegraphics[width=0.85\textwidth]{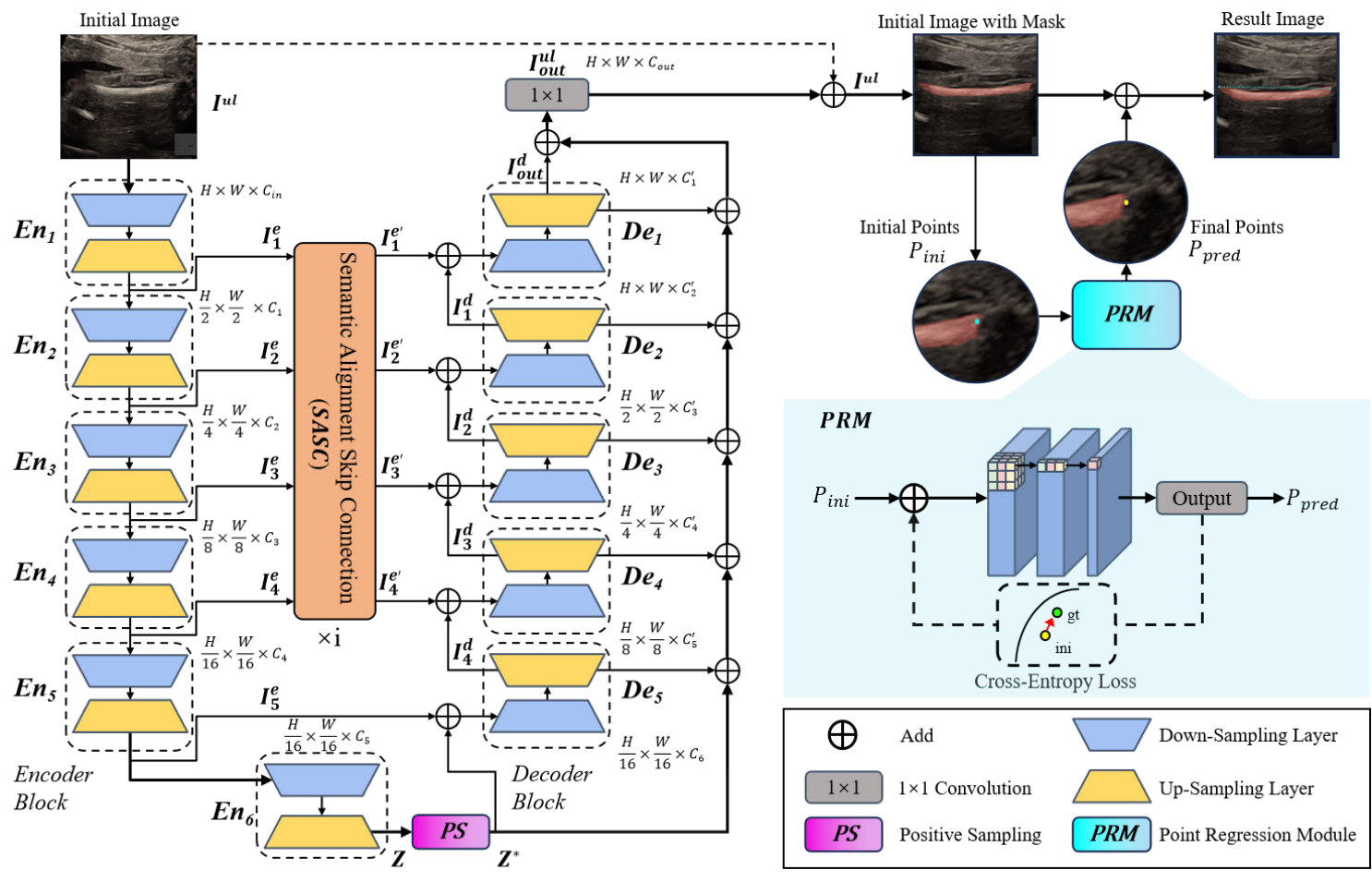}
\caption{The overall architecture of UniFLM. The backbone consists of a 6-stage Encoder ($\mathbf{En}$) and a 5-stage Decoder ($\mathbf{De}$). (1) \textbf{SASC}: Multi-scale encoder features $\mathbf{I}^e$ are projected and aligned via the centralized SASC module. (2) \textbf{PoSamp}: The Positive Sampling module at the bottleneck filters noise from feature $\mathbf{Z}$ to generate $\mathbf{Z}^*$. (3) \textbf{Segmentation \& PRM}: The decoder outputs an initial mask; initial points $\mathbf{P}_{ini}$ are extracted and fed into the PRM module to predict the final measurement points $\mathbf{P}_{pred}$.}
\label{fig:main_framework}
\end{figure}

The proposed \textbf{UniFLM} framework, illustrated in Fig.~\ref{fig:main_framework}, is designed for unified fetal long-bone segmentation and precise biometric measurement. UniFLM employs a deep 6-stage encoder ($\mathbf{En}_1$ to $\mathbf{En}_6$) and a 5-stage decoder ($\mathbf{De}_1$ to $\mathbf{De}_5$) based on the U2-Net architecture \cite{qin2020u2net}, with nested RSU blocks for multi-scale feature extraction.

The framework integrates three core modules:

\noindent\textbf{Semantic Alignment Skip Connection (SASC):} A centralized module that aggregates multi-scale encoder features ($\mathbf{I}^e_1 \dots \mathbf{I}^e_4$), aligns them via dual cross-attention mechanisms (Channel Cross-Attention and Spatial Cross-Attention), and distributes them ($\mathbf{I}^{e'}_1 \dots \mathbf{I}^{e'}_4$) to the corresponding decoder stages.

\noindent\textbf{Positive Sampling (PoSamp):} A bottleneck module that adaptively filters background noise from the deepest encoder feature $\mathbf{Z}$ using a learned gating mechanism, producing a robust representation $\mathbf{Z}^*$ for subsequent decoding.

\noindent\textbf{Point Regression Mapping (PRM):} A coarse-to-fine measurement head that refines initial keypoints $\mathbf{P}_{ini}$ derived from segmentation masks into precise landmarks $\mathbf{P}_{pred}$ using a dedicated ResNet50-based refinement network.

\vspace{0.5em}
\noindent\textbf{Multi-task Training Objective:}
The overall training objective combines segmentation and keypoint regression losses with explicit weights:
\begin{equation}
\mathcal{L}_{total} = \lambda_1 \cdot \mathcal{L}_{seg} + \lambda_2 \cdot \mathcal{L}_{kpt}
\label{eq:total_loss}
\end{equation}
where the weights $\lambda_1 = 1.0$ and $\lambda_2 = 0.5$ are determined via grid search on the validation set (see Section~\ref{subsec:sensitivity}).

The segmentation loss $\mathcal{L}_{seg}$ comprises Cross-Entropy and Dice losses:
\begin{equation}
\mathcal{L}_{seg} = \mathcal{L}_{CE}(\hat{\mathbf{M}}, \mathbf{M}_{gt}) + \mathcal{L}_{Dice}(\hat{\mathbf{M}}, \mathbf{M}_{gt})
\label{eq:seg_loss}
\end{equation}
\begin{equation}
\mathcal{L}_{CE} = -\sum_{i} \mathbf{M}_{gt}^{(i)} \log(\hat{\mathbf{M}}^{(i)})
\end{equation}
\begin{equation}
\mathcal{L}_{Dice} = 1 - \frac{2|\hat{\mathbf{M}} \cap \mathbf{M}_{gt}|}{|\hat{\mathbf{M}}| + |\mathbf{M}_{gt}|}
\end{equation}

The keypoint regression loss $\mathcal{L}_{kpt}$ uses Smooth L1:
\begin{equation}
\mathcal{L}_{kpt} = \sum_{j=1}^{4} \text{SmoothL1}(\hat{\mathbf{K}}_j - \mathbf{K}_{gt,j})
\label{eq:kpt_loss}
\end{equation}
\begin{equation}
\text{SmoothL1}(x) = \begin{cases} 0.5x^2 & \text{if } |x| < 1 \\ |x| - 0.5 & \text{otherwise} \end{cases}
\end{equation}
where $\hat{\mathbf{K}} = [\hat{x}_1, \hat{y}_1, \hat{x}_2, \hat{y}_2]$ denotes the predicted endpoint coordinates and $\mathbf{K}_{gt}$ the corresponding ground truth.

\begin{figure}[H]
\centering
\includegraphics[width=0.85\textwidth]{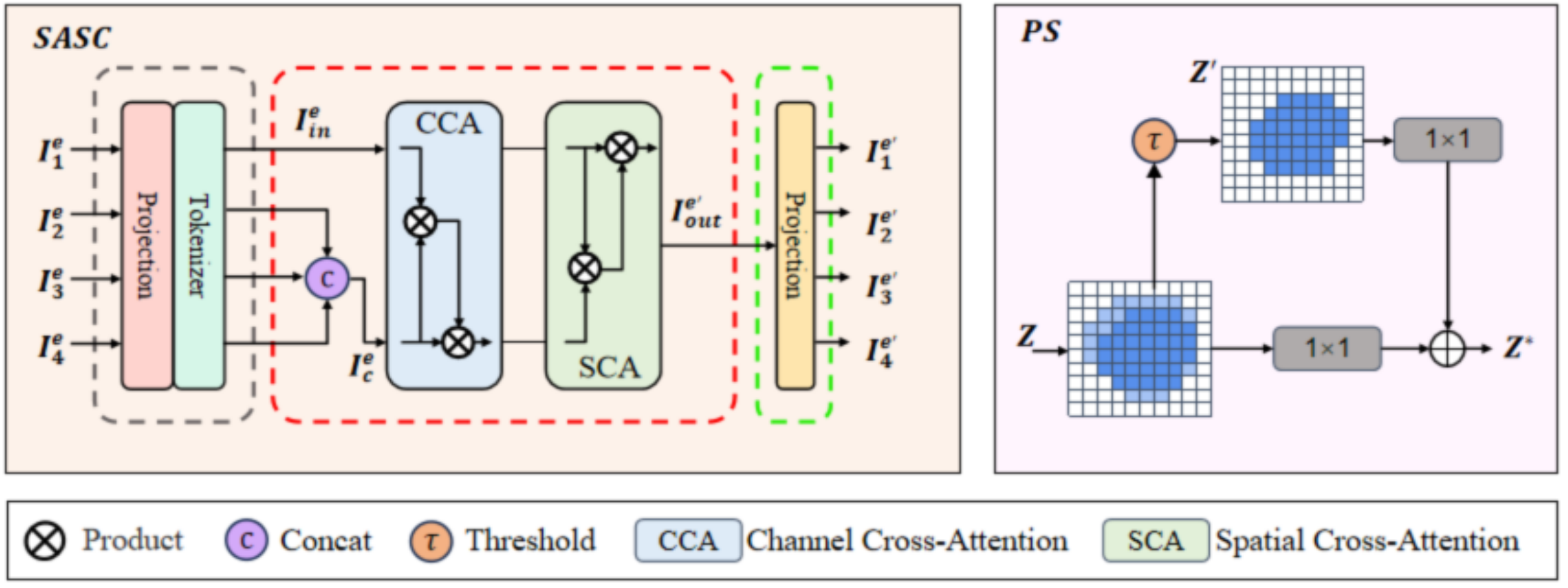}
\caption{Detailed structure of the proposed modules. \textbf{Left (SASC):} Multi-scale features $\mathbf{I}^e_i$ are projected and tokenized, concatenated into $\mathbf{I}^e_{c}$, then refined via Channel Cross-Attention (CCA) and Spatial Cross-Attention (SCA) before re-projection to original resolutions. \textbf{Right (PoSamp):} The deepest feature $\mathbf{Z}$ is filtered by a learned gating mechanism to produce $\mathbf{Z}^*$ via residual connection.}
\label{fig:modules_detail}
\end{figure}

\subsection{Inference Pipeline}
\label{sec:pseudocode}

The complete inference procedure is summarized in Algorithm~\ref{alg:uniflm}, with post-processing in Algorithm~\ref{alg:postprocess}.

\begin{algorithm}[htbp]
\caption{UniFLM Inference Pipeline}
\label{alg:uniflm}
\begin{algorithmic}[1]
\REQUIRE Ultrasound image $\mathbf{I} \in \mathbb{R}^{H \times W \times 3}$
\ENSURE Segmentation mask $\hat{\mathbf{M}}$, Final keypoints $\hat{\mathbf{K}}_{final}$
\STATE // \textbf{Stage 1: Segmentation \& Initial Keypoint Extraction}
\STATE $\mathbf{I}_{384} \gets \text{Resize}(\mathbf{I}, 384 \times 384)$; $\mathbf{I}_{norm} \gets \text{Normalize}(\mathbf{I}_{384}, \boldsymbol{\mu}_u, \boldsymbol{\sigma}_u)$
\STATE $\{\mathbf{E}_1, \dots, \mathbf{E}_6\} \gets \text{U2Net\_Encoder}(\mathbf{I}_{norm})$
\STATE $\mathbf{E}_6 \gets \text{PoSamp}(\mathbf{E}_6)$; $\{\mathbf{E}_1, \dots, \mathbf{E}_4\} \gets \text{SASC}(\mathbf{E}_1, \dots, \mathbf{E}_4)$
\STATE $\{\mathbf{D}_5, \dots, \mathbf{D}_1\} \gets \text{U2Net\_Decoder}(\{\mathbf{E}_4, \mathbf{E}_5, \mathbf{E}_6\})$
\STATE $\hat{\mathbf{M}}_{raw} \gets \text{Argmax}(\text{Conv}_{1\times1}(\text{Concat}(\text{Side}(\mathbf{D}_1, \dots, \mathbf{D}_6))))$
\STATE $\hat{\mathbf{M}} \gets \text{PostProcess}(\hat{\mathbf{M}}_{raw})$\hfill $\triangleright$ Alg.~\ref{alg:postprocess}
\STATE $\mathbf{K}_{init} \gets \text{SortLR}(\text{FarthestPair}(\text{FindContour}(\hat{\mathbf{M}})))$
\STATE // \textbf{Stage 2: Keypoint Refinement (PRM)}
\STATE $\mathbf{I}_{256} \gets \text{Resize}(\mathbf{I}, 256 \times 192)$
\STATE $\boldsymbol{\Delta} \gets \text{ResNet50}(\text{Normalize}(\mathbf{I}_{256}, \boldsymbol{\mu}_p, \boldsymbol{\sigma}_p))$\hfill $\triangleright$ 4-dim offset
\STATE $\hat{\mathbf{K}}_{final} \gets \mathbf{K}_{init} + \boldsymbol{\Delta}$
\RETURN $\hat{\mathbf{M}}, \hat{\mathbf{K}}_{final}$
\end{algorithmic}
\end{algorithm}

\begin{algorithm}[htbp]
\caption{Post-Processing: Swap \& Check}
\label{alg:postprocess}
\begin{algorithmic}[1]
\REQUIRE Multi-class mask $\mathbf{M} \in \{0,1,2\}^{H \times W}$
\ENSURE Cleaned mask $\mathbf{M}_{clean}$
\STATE $\mathbf{M} \gets \text{MorphologicalOpen}(\mathbf{M})$ \COMMENT{Remove noise speckles}
\STATE $\mathbf{M} \gets \text{MorphologicalClose}(\mathbf{M})$ \COMMENT{Fill internal holes}
\STATE $\mathbf{M} \gets \text{KeepLargestComponents}(\mathbf{M}, k=2)$ \COMMENT{Retain top-2 CCs}
\FOR{each class $c \in \{1, 2\}$}
    \STATE Retain only the largest connected component of $(\mathbf{M} == c)$
\ENDFOR
\FOR{each connected component $CC$ in $\mathbf{M}$}
    \IF{$CC.\text{label} == 1$ and $CC$ predominantly overlaps with label-2 bone region}
        \STATE Swap $CC$ to label 2
    \ELSIF{$CC.\text{label} == 2$ and $CC$ predominantly overlaps with label-1 bone region}
        \STATE Swap $CC$ to label 1
    \ENDIF
\ENDFOR
\RETURN $\mathbf{M}_{clean}$
\end{algorithmic}
\end{algorithm}

\subsection{Semantic Alignment Skip Connection (SASC)}
\label{sec:sasc}

Standard skip connections often fail to handle the semantic discrepancy between shallow encoder features (rich in texture but noisy) and deep decoder features. To address this, as shown in Fig.~\ref{fig:modules_detail} (Left), our SASC module acts as a global feature aligner, bridging the semantic gap before fusing representations into the decoder.

Let $\{\mathbf{I}^e_1, \mathbf{I}^e_2, \mathbf{I}^e_3, \mathbf{I}^e_4\}$ denote the feature maps from the first four encoder blocks. These multi-scale features are projected into a unified embedding space via a Projection layer and Tokenizer, then concatenated:
\begin{equation}
\mathbf{I}^e_{c} = \text{Concat}(\text{Tokenizer}(\text{Proj}(\mathbf{I}^e_i))), \quad i \in \{1, 2, 3, 4\}
\label{eq:concat}
\end{equation}

The unified feature $\mathbf{I}^e_{c}$ is processed by a dual-attention mechanism:

\noindent\textbf{Channel Cross-Attention (CCA):} Captures inter-channel dependencies via scaled dot-product attention over channel dimension, enhancing structurally relevant feature maps:
\begin{equation}
\mathbf{A}_c = \text{softmax}\left(\frac{\mathbf{Q}_c \cdot \mathbf{K}_c^T}{\sqrt{d_c}}\right) \cdot \mathbf{V}_c
\end{equation}

\noindent\textbf{Spatial Cross-Attention (SCA):} Models long-range spatial dependencies to distinguish bone structures from acoustic shadows and artifacts:
\begin{equation}
\mathbf{A}_s = \text{softmax}\left(\frac{\mathbf{Q}_s^T \cdot \mathbf{K}_s}{\sqrt{d_s}}\right) \cdot \mathbf{V}_s
\end{equation}

The refined output is produced as:
\begin{equation}
\mathbf{I}^{e'}_{out} = \text{SCA}(\text{CCA}(\mathbf{I}^e_{c}))
\end{equation}

Finally, a reverse projection layer redistributes the features back to their original spatial resolutions, yielding aligned features $\{\mathbf{I}^{e'}_1, \dots, \mathbf{I}^{e'}_4\}$, which are fused with decoder features via element-wise summation.

\subsection{Positive Sampling (PoSamp) Module}
\label{sec:ps}

Ultrasound images inherently suffer from low signal-to-noise ratio and complex acoustic artifacts. To prevent noise propagation from encoder to decoder, we introduce the PoSamp module at the bottleneck (Fig.~\ref{fig:modules_detail} Right).

Taking the deepest encoder feature $\mathbf{Z}$ from $\mathbf{En}_6$ as input, the PoSamp module applies a gating mechanism. The feature is first normalized and pooled, then a learned gate $\mathbf{G}$ controls information flow:
\begin{equation}
\mathbf{G} = \sigma(\mathbf{W}_g \cdot \text{AvgPool}(\text{LayerNorm}(\mathbf{Z})) + \mathbf{b}_g)
\label{eq:gate}
\end{equation}
\begin{equation}
\mathbf{Z}' = \text{Conv}_{1\times1}(\mathbf{Z} \odot \mathbf{G})
\label{eq:gated}
\end{equation}

where $\sigma$ is the sigmoid activation and $\odot$ denotes Hadamard product. A residual connection preserves structural integrity:
\begin{equation}
\mathbf{Z}^* = \text{Conv}_{1\times1}(\mathbf{Z}) + \mathbf{Z}'
\label{eq:residual_z}
\end{equation}

The output $\mathbf{Z}^*$ serves as the clean, semantically rich input for the first decoder stage $\mathbf{De}_5$, mitigating the effects of inherent ultrasound acoustic artifacts.

\subsection{Point Regression Mapping (PRM)}
\label{sec:prm}

To achieve precise biometric measurement, we propose a coarse-to-fine PRM strategy (Fig.~\ref{fig:main_framework} Right).

\subsubsection{Initial Point Extraction}
The decoder generates a segmentation probability map $\mathbf{I}^{ul}$. Post-processing extracts the bone contour, from which the two farthest points (by $\ell_2$ distance) are selected as initial keypoints $\mathbf{P}_{ini}$. These are then sorted left-to-right by $x$-coordinate (LR sorting), ensuring consistent ordering for subsequent refinement.

\subsubsection{Patch-based Refinement}
$\mathbf{P}_{ini}$ may be inaccurate due to boundary ambiguity. PRM uses a ResNet50 backbone to predict coordinate offsets $\boldsymbol{\Delta}$ from the image and the initial keypoint context:
\begin{equation}
\hat{\mathbf{K}}_{final} = \mathbf{K}_{init} + \boldsymbol{\Delta}
\label{eq:prm_offset}
\end{equation}

The refinement is supervised via a Cross-Entropy loss treating landmark localization as spatial classification, which improves convergence stability compared to direct coordinate regression:
\begin{equation}
\mathcal{L}_{PRM} = \text{CrossEntropy}(\mathbf{P}_{pred}, \mathbf{P}_{gt})
\label{eq:prm_loss}
\end{equation}
where $\mathbf{P}_{pred}$ is the predicted probability heatmap of the landmark location, and $\mathbf{P}_{gt}$ is the corresponding ground truth coordinate annotated by senior clinicians.

\section{Experimental Results}
\label{sec:experiments}

\subsection{Dataset Construction and Statistics}
\label{subsec:dataset}

\subsubsection{Data Collection}

We established the Fetal Limb Bones (FLB) dataset through collaboration with multiple clinical centers, collecting ultrasound images acquired between 2017 and 2023. The dataset comprises 1,690 images covering four anatomical categories: 600 images of the humerus, 500 images of the femur, 295 images of the forearm (radius-ulna), and 295 images of the leg (tibia-fibula).

Images were acquired using various ultrasound systems (GE Voluson, Philips EPIQ, Samsung) across gestational ages ranging from 14 to 40 weeks, ensuring diversity in image quality and fetal growth development stages. Each image captures a standard diagnostic plane for the corresponding long bone, with varying probe positions and scanning angles depending on the target anatomy.

\subsubsection{Annotation Protocol}

All images were annotated by three experienced sonographers ($>$10 years experience) following ISUOG guidelines \cite{salomon2019isuog}. For each image, annotators provided:
\begin{enumerate}[label=(\arabic*), leftmargin=*]
\item Pixel-level segmentation mask delineating the bone boundaries.
\item Endpoint coordinates marking the proximal and distal bone margins.
\item Quality assessment score (1--5) indicating image clarity.
\end{enumerate}

Inter-annotator agreement was assessed using Dice coefficient (mean: 0.92) and endpoint distance (mean: 1.1 mm), demonstrating high consistency. Final annotations were derived through a majority voting protocol, supplemented by senior expert adjudication to resolve persistent disagreements.

\subsubsection{Dataset Split}

We adopted a patient-wise split strategy (7:1:2 for training/validation/test) to prevent data leakage. Images from the same patient appear exclusively in one subset, providing realistic evaluation of generalization to unseen clinical cases.

\begin{figure}[H]
\centering
\includegraphics[width=\textwidth]{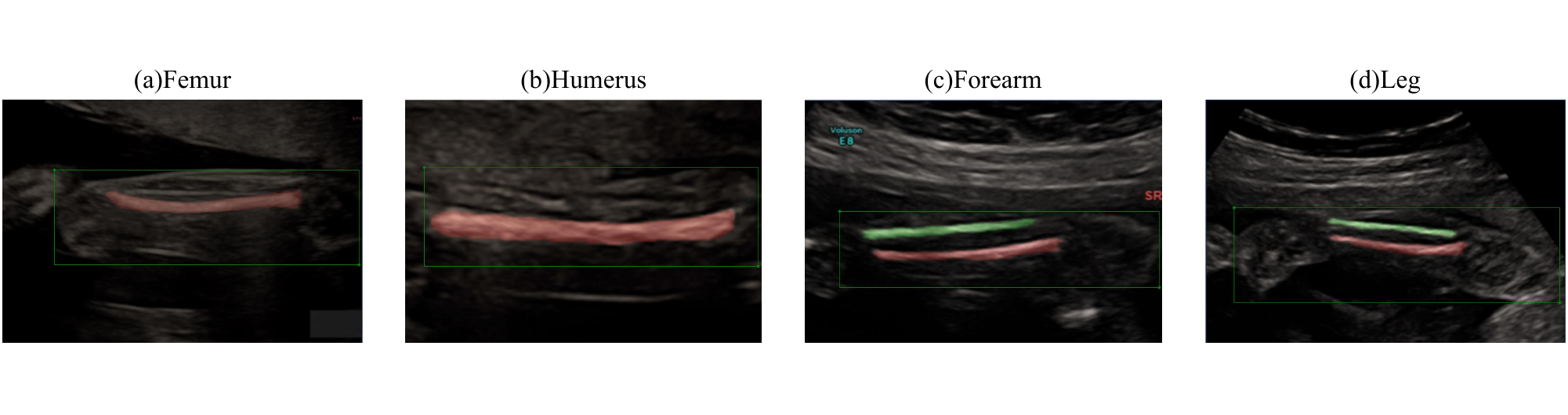}
\caption{Representative samples from the FLB dataset showing (a) Femur, (b) Humerus, (c) Forearm, and (d) Leg. Red contours denote expert annotations.}
\label{fig:dataset_examples}
\end{figure}

\begin{figure}[H]
    \centering
    \includegraphics[width=0.6\textwidth]{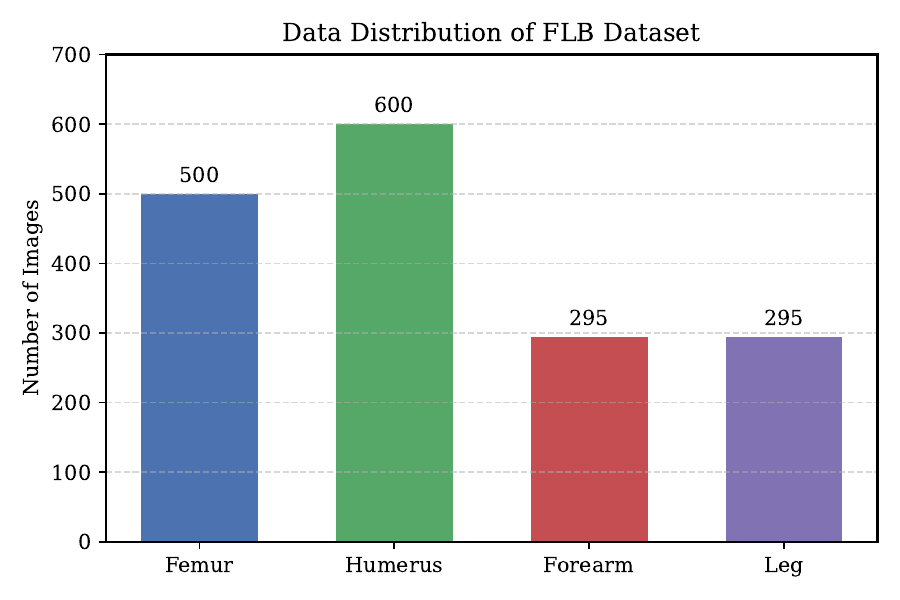}
    \caption{Distribution of samples across the four bone categories in the FLB dataset, showing gestational age distribution within each category.}
    \label{fig:dataset_hist}
\end{figure}

\subsection{Implementation Details}
\label{subsec:implementation}

\subsubsection{Experimental Setup}

All experiments were conducted using PyTorch 1.12 on a single NVIDIA Tesla V100 GPU (32GB memory). Images were resized to $256 \times 256$ pixels with bilinear interpolation. The batch size was set to 8, with a standard weight decay of $1 \times 10^{-4}$. The initial learning rate was set to $2 \times 10^{-2}$ for SGD and $1 \times 10^{-4}$ for AdamW. Loss weights were $\lambda_1 = 1.0$ and $\lambda_2 = 0.5$ (Eq.~\ref{eq:total_loss}). For reproducibility, all experiments were repeated with 3 random seeds (42, 123, 456), and results are reported as mean $\pm$ standard deviation across seeds.

\subsubsection{Evaluation Metrics}

For segmentation evaluation, we employ Dice and Intersection over Union (IoU) for regional overlap, and Hausdorff Distance (HD95) for spatial boundary accuracy. For measurement evaluation, we use Mean Endpoint Distance (MED), Mean Absolute Error (MAE), Mean Squared Error (MSE), Root Mean Square Error (RMSE), and the Distance-based Point Endpoint Distance Error (DPEDE), defined as the sum of endpoint Euclidean distances between predicted and ground truth keypoints. We also report the coefficient of determination ($R^2$) and Normalized Mean Error (NME) for cross-dataset comparison.

\textbf{Statistical reporting.} Following ISUOG guidelines for fetal biometry validation \cite{papageorghiou2017quality}, all metrics are reported as mean $\pm$ standard deviation across test samples. We additionally compute 95\% confidence intervals via bootstrap (1,000 iterations) and Bland-Altman agreement analysis \cite{Bland1986} to quantify systematic vs.\ random error. Throughout the paper, $\sigma$ denotes sigmoid activation; $\odot$ denotes Hadamard (element-wise) product; $\text{Conv}_{1\times1}$ denotes $1\times1$ convolution; $\text{AvgPool}$ denotes global average pooling; and $\text{LayerNorm}$ denotes layer normalization.

\subsection{Comparative Analysis}
\label{subsec:comparison}

\subsubsection{Baseline Methods}

We compared UniFLM against: (a) \textbf{Classic architectures}: U-Net \cite{Ronneberger2015Unet}, UNet++ \cite{unetplusplus}, Attention U-Net \cite{Oktay2018}; (b) \textbf{Transformer-based}: TransUNet \cite{transunet}, Swin-Unet \cite{swinunet}; (c) \textbf{Foundation models}: MedSAM \cite{Ma2024MedSAM}, SAM-US \cite{lin2023samus}; (d) \textbf{Recent advances}: VM-UNet \cite{Ruan2024VMUNet}, U-KAN \cite{Li2024UKAN}. All baselines were trained using official implementations with hyperparameters tuned on our validation set.

\subsubsection{Quantitative Results}

Table~\ref{tab:comparison} presents comprehensive quantitative comparison. Results are reported as Mean $\pm$ SD across test samples.

\begin{table}[H]
\centering
\caption{Quantitative comparison on the FLB dataset (Mean $\pm$ SD across test samples). \textbf{Bold}: best; \underline{Underline}: second best.}
\label{tab:comparison}
\small
\setlength{\tabcolsep}{2.5pt}
\begin{tabular}{lcccc}
\toprule
\multirow{2}{*}{\textbf{Model}} & \multicolumn{2}{c}{\textbf{Femur}} & \multicolumn{2}{c}{\textbf{Humerus}} \\
\cmidrule(lr){2-3} \cmidrule(lr){4-5}
& Dice (\%) & IoU (\%) & Dice (\%) & IoU (\%) \\
\midrule
UNet \cite{Ronneberger2015Unet} & $82.66_{\pm4.21}$ & $72.49_{\pm5.11}$ & $87.06_{\pm3.87}$ & $78.92_{\pm4.45}$ \\
UNet++ \cite{unetplusplus} & $88.56_{\pm2.15}$ & $80.88_{\pm2.76}$ & $89.43_{\pm1.98}$ & $81.61_{\pm2.55}$ \\
Attention U-Net \cite{Oktay2018} & $87.21_{\pm2.54}$ & $79.15_{\pm3.12}$ & $88.76_{\pm2.31}$ & $80.42_{\pm2.95}$ \\
SwinUNet \cite{swinunet} & $78.99_{\pm5.42}$ & $67.90_{\pm6.78}$ & $78.99_{\pm5.01}$ & $67.51_{\pm6.12}$ \\
TransUNet \cite{transunet} & $87.14_{\pm2.89}$ & $78.96_{\pm3.45}$ & $89.47_{\pm2.12}$ & $81.63_{\pm2.68}$ \\
MedSAM \cite{Ma2024MedSAM} & $86.50_{\pm3.10}$ & $78.10_{\pm3.85}$ & $88.20_{\pm2.65}$ & $80.15_{\pm3.22}$ \\
SAM-US \cite{lin2023samus} & $87.90_{\pm2.25}$ & $80.25_{\pm2.80}$ & $89.15_{\pm2.05}$ & $81.80_{\pm2.48}$ \\
VM-UNet \cite{Ruan2024VMUNet} & $\underline{88.65}_{\pm2.01}$ & $\underline{81.10}_{\pm2.55}$ & $\underline{89.50}_{\pm1.85}$ & $\underline{82.10}_{\pm2.22}$ \\
U-KAN \cite{Li2024UKAN} & $88.10_{\pm2.18}$ & $80.50_{\pm2.72}$ & $89.10_{\pm2.08}$ & $81.90_{\pm2.45}$ \\
\rowcolor{gray!10}
\textbf{UniFLM (Ours)} & $\mathbf{89.90}_{\pm1.87}$ & $\mathbf{82.19}_{\pm2.38}$ & $\mathbf{90.76}_{\pm1.65}$ & $\mathbf{83.83}_{\pm2.10}$ \\
\bottomrule
\multicolumn{5}{c}{}\\[-6pt]
\multirow{2}{*}{\textbf{Model}} & \multicolumn{2}{c}{\textbf{Forearm}} & \multicolumn{2}{c}{\textbf{Leg}} \\
\cmidrule(lr){2-3} \cmidrule(lr){4-5}
& Dice (\%) & IoU (\%) & Dice (\%) & IoU (\%) \\
\midrule
UNet \cite{Ronneberger2015Unet} & $66.15_{\pm6.32}$ & $52.97_{\pm7.45}$ & $71.64_{\pm5.78}$ & $58.93_{\pm6.85}$ \\
UNet++ \cite{unetplusplus} & $66.96_{\pm5.89}$ & $53.56_{\pm6.91}$ & $71.64_{\pm5.42}$ & $58.37_{\pm6.55}$ \\
Attention U-Net \cite{Oktay2018} & $67.23_{\pm5.45}$ & $54.12_{\pm6.52}$ & $72.18_{\pm5.18}$ & $59.45_{\pm6.32}$ \\
SwinUNet \cite{swinunet} & $49.00_{\pm9.12}$ & $36.59_{\pm10.21}$ & $51.05_{\pm8.95}$ & $38.11_{\pm9.85}$ \\
TransUNet \cite{transunet} & $62.79_{\pm6.78}$ & $50.17_{\pm7.56}$ & $70.24_{\pm5.98}$ & $57.12_{\pm6.72}$ \\
MedSAM \cite{Ma2024MedSAM} & $65.40_{\pm6.25}$ & $51.80_{\pm7.15}$ & $72.10_{\pm5.45}$ & $59.20_{\pm6.28}$ \\
SAM-US \cite{lin2023samus} & $67.80_{\pm5.55}$ & $54.10_{\pm6.45}$ & $73.05_{\pm5.12}$ & $60.15_{\pm5.95}$ \\
VM-UNet \cite{Ruan2024VMUNet} & $\underline{68.20}_{\pm5.32}$ & $54.80_{\pm6.15}$ & $73.15_{\pm4.98}$ & $60.50_{\pm5.78}$ \\
U-KAN \cite{Li2024UKAN} & $68.10_{\pm5.48}$ & $\underline{55.90}_{\pm6.08}$ & $\underline{73.80}_{\pm4.85}$ & $\underline{61.20}_{\pm5.55}$ \\
\rowcolor{gray!10}
\textbf{UniFLM (Ours)} & $\mathbf{70.55}_{\pm4.85}$ & $\mathbf{57.12}_{\pm5.65}$ & $\mathbf{74.50}_{\pm4.42}$ & $\mathbf{62.35}_{\pm5.18}$ \\
\bottomrule
\end{tabular}
\end{table}

UniFLM achieves the best overall performance across all categories. On single-bone structures (Femur, Humerus), it achieves $1.25\%$ and $1.26\%$ Dice improvements over the best baseline (VM-UNet). Improvements are more pronounced on paired-bone structures (Forearm +$2.35\%$, Leg +$0.70\%$ Dice), where semantic gaps and class confusion pose greater challenges. The reported standard deviations, consistently lower for UniFLM than baselines, indicate superior stability across diverse test samples.

\subsubsection{Statistical Significance}

To validate that observed improvements are statistically significant, we conducted paired $t$-tests comparing UniFLM against key baselines (Table~\ref{tab:p_values}).

\begin{table}[H]
\centering
\caption{Statistical significance ($p$-values from paired $t$-tests). Values $< 0.05$ (bold) indicate significance at the 95\% confidence level.}
\label{tab:p_values}
\small
\begin{tabular}{lcccc}
\toprule
\textbf{Comparison} & \textbf{Femur} & \textbf{Humerus} & \textbf{Forearm} & \textbf{Leg} \\
\midrule
UniFLM vs. UNet & \textbf{$<$0.001} & \textbf{$<$0.001} & \textbf{$<$0.001} & \textbf{$<$0.001} \\
UniFLM vs. UNet++ & 0.087 & 0.124 & \textbf{0.002} & \textbf{0.012} \\
UniFLM vs. SAM-US & \textbf{0.035} & \textbf{0.041} & \textbf{$<$0.001} & \textbf{0.004} \\
UniFLM vs. VM-UNet & 0.156 & 0.203 & \textbf{0.028} & \textbf{0.033} \\
UniFLM vs. U-KAN & 0.092 & 0.118 & \textbf{0.015} & \textbf{0.041} \\
\bottomrule
\end{tabular}
\end{table}

UniFLM shows statistically significant improvements ($p < 0.05$) over all baselines on the more challenging Forearm and Leg datasets. On Femur and Humerus, improvements are consistent in direction but do not always reach $p<0.05$ against the strongest baselines due to high initial performance. Overall, the results support the effectiveness of the SASC and PoSamp modules, particularly for complex multi-bone structures.

\subsection{Ablation Study}
\label{subsec:ablation}

\subsubsection{Module Contribution Analysis}

We systematically evaluate each module's contribution through ablation experiments (Table~\ref{tab:ablation}).

\begin{table}[H]
\centering
\caption{Ablation study on module effectiveness (Mean Dice $\pm$ SD across 3 seeds).}
\label{tab:ablation}
\small
\setlength{\tabcolsep}{2pt}
\begin{tabular}{ccc*{4}{c}}
\toprule
\multicolumn{3}{c}{\textbf{Module}} & \multicolumn{4}{c}{\textbf{Dice (\%)}} \\
\cmidrule(r){1-3} \cmidrule(r){4-7}
SASC & PoSamp & PRM & Femur & Humerus & Forearm & Leg \\
\midrule
\xmark & \xmark & \xmark & $86.96_{\pm0.52}$ & $89.43_{\pm0.45}$ & $68.36_{\pm0.78}$ & $73.69_{\pm0.65}$ \\
\cmark & \xmark & \xmark & $87.82_{\pm0.48}$ & $89.36_{\pm0.42}$ & $67.02_{\pm0.82}$ & $73.05_{\pm0.68}$ \\
\xmark & \cmark & \xmark & $87.55_{\pm0.50}$ & $89.57_{\pm0.40}$ & $68.53_{\pm0.75}$ & $72.21_{\pm0.70}$ \\
\cmark & \cmark & \xmark & $88.70_{\pm0.42}$ & $90.06_{\pm0.35}$ & $69.55_{\pm0.68}$ & $73.80_{\pm0.58}$ \\
\rowcolor{gray!10}
\cmark & \cmark & \cmark & $\mathbf{89.90}_{\pm0.38}$ & $\mathbf{90.76}_{\pm0.32}$ & $\mathbf{70.55}_{\pm0.62}$ & $\mathbf{74.50}_{\pm0.52}$ \\
\bottomrule
\end{tabular}
\end{table}

Key observations: (1) SASC alone yields $+0.86\%$ Dice on Femur but degrades Forearm by $-1.34\%$, suggesting semantic alignment is most beneficial when paired with robust feature supervision. (2) PoSamp alone provides consistent improvements across all categories ($+0.14\%$ Dice on Humerus). (3) Combining SASC and PoSamp yields synergistic gains exceeding the sum of individual contributions, particularly on Forearm ($+1.19\%$ over baseline). (4) PRM adds further improvements, notably on Leg ($+0.70\%$ Dice).

\subsubsection{Measurement Module Analysis}

Table~\ref{tab:measurement_ablation} evaluates PRM's impact on measurement accuracy.

\begin{table}[H]
\centering
\caption{Impact of PRM on measurement accuracy (Mean $\pm$ SD). MED: Mean Endpoint Distance (px), MAE: Mean Absolute Error (mm).}
\label{tab:measurement_ablation}
\footnotesize
\setlength{\tabcolsep}{3.5pt}
\begin{tabular}{lcccc}
\toprule
\textbf{Method} & \textbf{Femur} & \textbf{Humerus} & \textbf{Forearm} & \textbf{Leg} \\
& {MED / MAE} & {MED / MAE} & {MED / MAE} & {MED / MAE} \\
\midrule
Geometric (skeleton) & $4.21_{\pm1.2}$ / $1.85_{\pm0.5}$ & $3.98_{\pm1.1}$ / $1.72_{\pm0.4}$ & $6.54_{\pm1.8}$ / $2.89_{\pm0.7}$ & $7.12_{\pm2.0}$ / $3.15_{\pm0.8}$ \\
Geometric (ellipse)  & $3.87_{\pm1.0}$ / $1.68_{\pm0.4}$ & $3.65_{\pm0.9}$ / $1.58_{\pm0.4}$ & $5.98_{\pm1.6}$ / $2.64_{\pm0.6}$ & $6.45_{\pm1.8}$ / $2.85_{\pm0.7}$ \\
Regression (direct)  & $3.45_{\pm0.9}$ / $1.52_{\pm0.4}$ & $3.21_{\pm0.8}$ / $1.41_{\pm0.3}$ & $5.12_{\pm1.4}$ / $2.26_{\pm0.6}$ & $5.78_{\pm1.6}$ / $2.55_{\pm0.6}$ \\
\rowcolor{gray!10}
\textbf{PRM (ours)}     & $\mathbf{2.89}_{\pm0.7}$ / $\mathbf{1.27}_{\pm0.3}$ & $\mathbf{2.76}_{\pm0.6}$ / $\mathbf{1.21}_{\pm0.3}$ & $\mathbf{4.38}_{\pm1.2}$ / $\mathbf{1.93}_{\pm0.5}$ & $\mathbf{4.91}_{\pm1.4}$ / $\mathbf{2.17}_{\pm0.5}$ \\
\bottomrule
\end{tabular}
\end{table}

PRM reduces MED by 16--23\% and MAE by 15--18\% compared to geometric methods, demonstrating that learning clinician-style measurement priors substantially improves anatomical landmark localization precision.

\subsection{Clinical Reliability Analysis}
\label{subsec:reliability}

\subsubsection{Error Distribution Analysis}

\begin{figure}[H]
\centering
\includegraphics[width=0.7\textwidth]{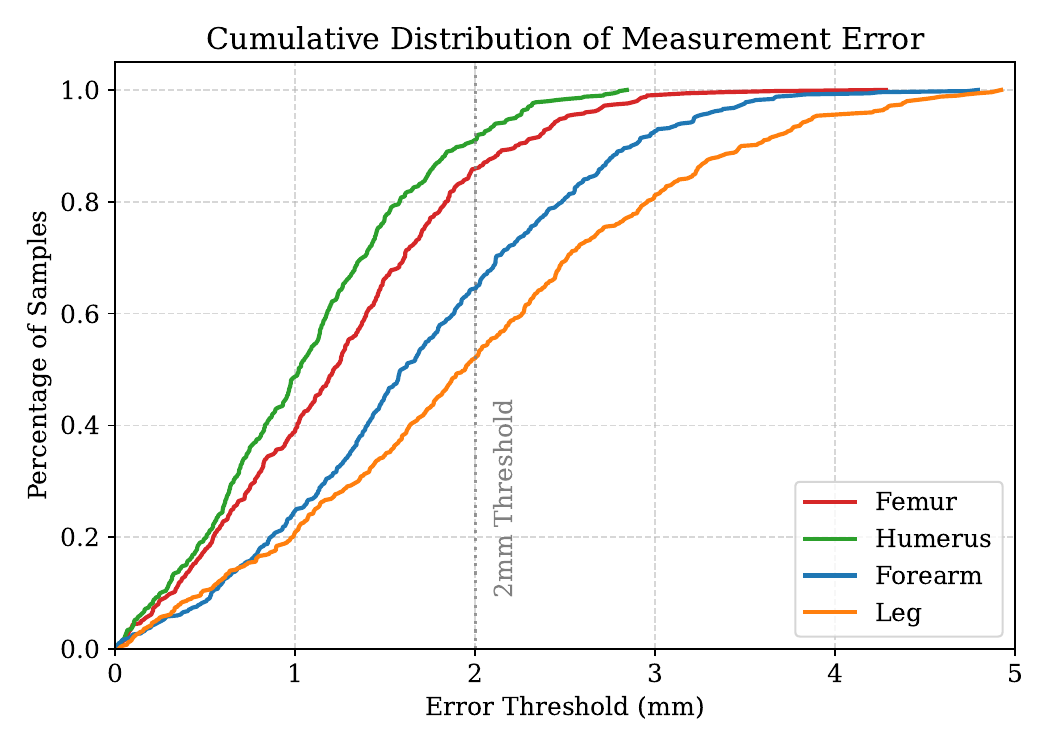}
\caption{Cumulative Distribution Function (CDF) of measurement errors. The vertical dashed line indicates the clinically acceptable threshold of 2.0 mm.}
\label{fig:cdf_curve}
\end{figure}

Figure~\ref{fig:cdf_curve} shows the Cumulative Distribution Function (CDF) of measurement errors. For Femur and Humerus, over 85\% of measurements fall within the clinically acceptable threshold of 2.0 mm. For the more challenging Forearm and Leg, approximately 75\% meet this criterion, with 90\% within 3.0 mm.

\subsubsection{Gestational Age Analysis}

\begin{figure}[H]
\centering
\includegraphics[width=1.0\textwidth]{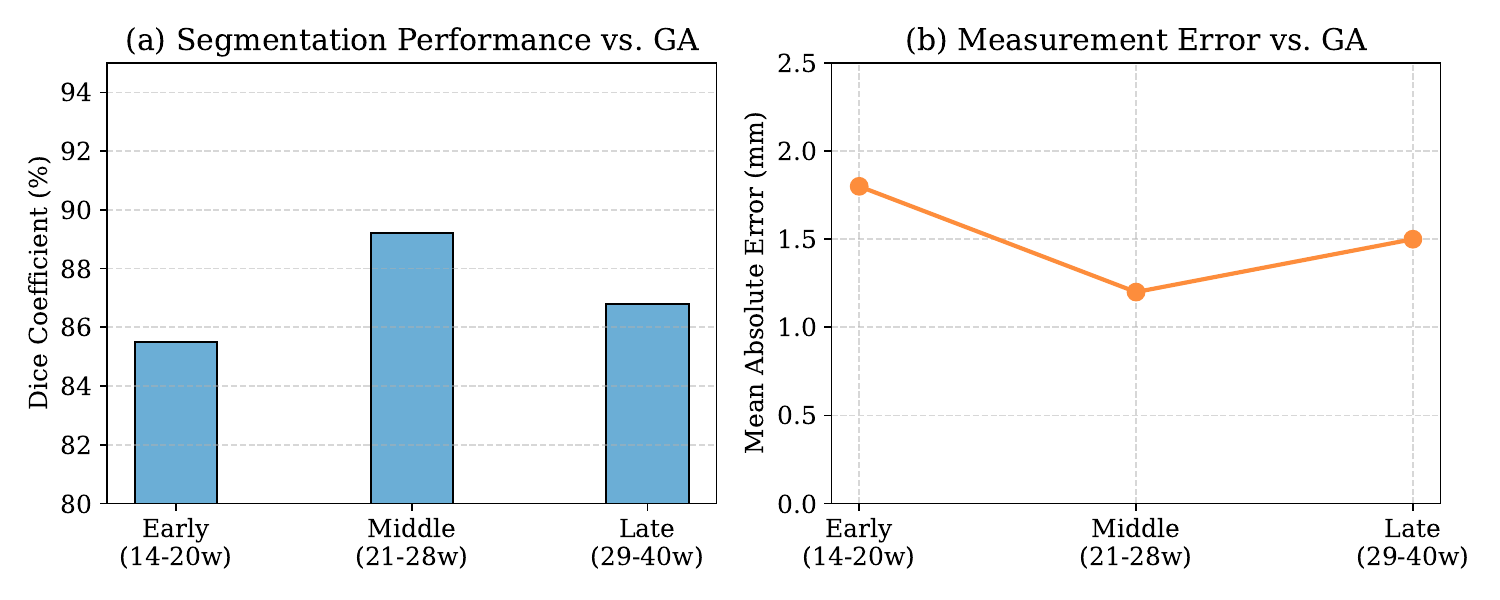}
\caption{Performance across gestational age groups. (a) Dice coefficient distribution. (b) Measurement error (MAE) distribution.}
\label{fig:ga_analysis}
\end{figure}

Performance stratified by gestational age (Fig.~\ref{fig:ga_analysis}) shows stability across Early (14--20 weeks), Middle (21--28 weeks), and Late (29--40 weeks) periods, with a slight decrease in Late GA due to increased acoustic shadowing from bone calcification.

\subsubsection{Robustness to Image Quality}

\begin{figure}[H]
\centering
\includegraphics[width=0.7\textwidth]{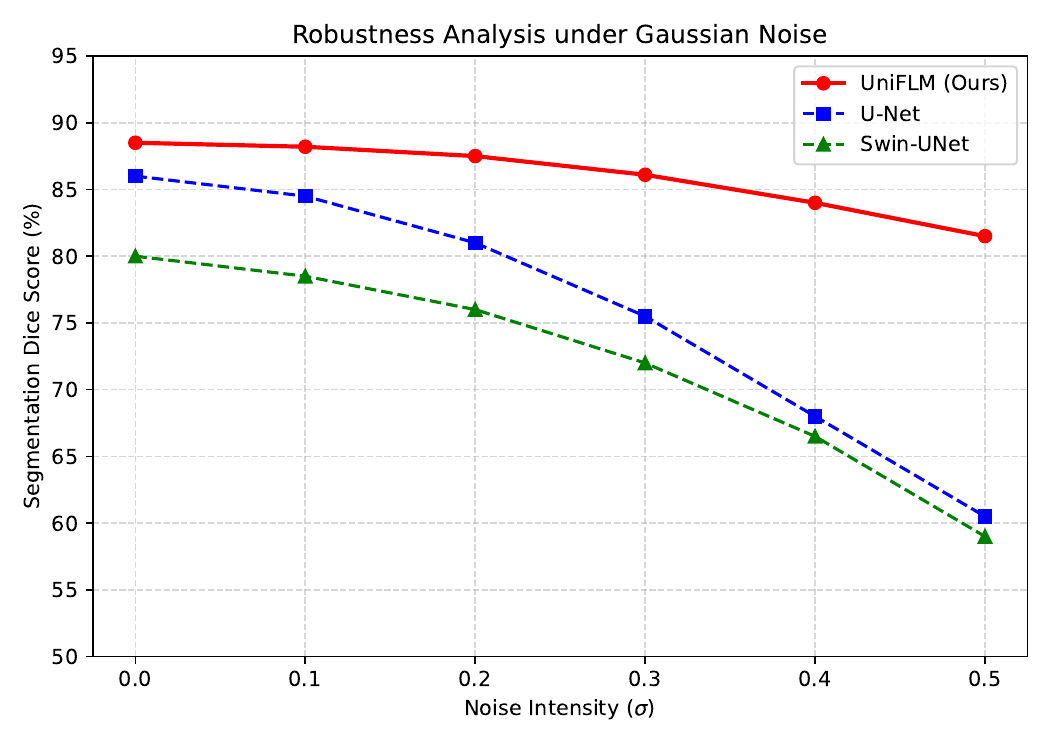}
\caption{Robustness analysis under varying noise levels.}
\label{fig:robustness}
\end{figure}

Under synthetic Gaussian noise (Fig.~\ref{fig:robustness}), UniFLM maintains Dice scores above 80\% at moderate noise levels ($\sigma \leq 0.1$), outperforming standard U-Net. This robustness is attributed to PoSamp's noise filtering and SASC's semantic alignment.

\subsection{Computational Efficiency}
\label{subsec:efficiency}

Table~\ref{tab:efficiency} compares computational characteristics.

\begin{table}[H]
\centering
\caption{Computational efficiency comparison (Input: $256 \times 256$, GPU: V100).}
\label{tab:efficiency}
\small
\begin{tabular}{lccccc}
\toprule
\textbf{Model} & \textbf{Params (M)} & \textbf{GFLOPs} & \textbf{FPS} & \textbf{Dice (\%)} & \textbf{MAE (mm)} \\
\midrule
UNet & \textbf{34.5} & \textbf{65.4} & \textbf{120} & 76.82 & 2.40 \\
UNet++ & 36.6 & 84.2 & 95 & 79.15 & 2.12 \\
TransUNet & 105.3 & 110.4 & 65 & 77.41 & 2.28 \\
MedSAM & 93.7 & 156.8 & 42 & 78.05 & 2.18 \\
VM-UNet & 42.1 & 78.5 & 88 & 79.88 & 2.05 \\
\rowcolor{gray!10}
\textbf{UniFLM} & 47.2 & 72.1 & 102 & \textbf{81.43} & \textbf{1.65} \\
\bottomrule
\end{tabular}
\end{table}

UniFLM achieves a favorable accuracy-efficiency trade-off with 47.2M parameters and 102 FPS, suitable for real-time clinical deployment on standard GPU hardware. The parameter count includes the U2-Net backbone ($\sim$45M), SASC (+0.53M), PoSamp (+0.46M), and PRM refinement head (+1.2M).

\subsection{Inter-observer Variability Comparison}
\label{subsec:interobserver}

Table~\ref{tab:inter_observer} compares UniFLM measurements against inter-observer variability among human experts.

\begin{table}[H]
\centering
\caption{Inter-observer variability analysis (measurement error in mm, Mean $\pm$ SD).}
\label{tab:inter_observer}
\small
\begin{tabular}{lcccc}
\toprule
\textbf{Comparison} & \textbf{Femur} & \textbf{Humerus} & \textbf{Forearm} & \textbf{Leg} \\
\midrule
Expert 1 vs. Expert 2 & $1.10 \pm 0.8$ & $0.95 \pm 0.6$ & $1.45 \pm 1.1$ & $1.52 \pm 1.2$ \\
Expert 1 vs. Expert 3 & $1.15 \pm 0.9$ & $1.02 \pm 0.7$ & $1.38 \pm 1.0$ & $1.48 \pm 1.1$ \\
Expert 2 vs. Expert 3 & $1.08 \pm 0.7$ & $0.98 \pm 0.6$ & $1.42 \pm 1.0$ & $1.55 \pm 1.2$ \\
\midrule
UniFLM vs. Expert 1 & $1.25 \pm 0.9$ & $1.18 \pm 0.7$ & $1.85 \pm 1.3$ & $2.05 \pm 1.4$ \\
UniFLM vs. Expert 2 & $1.28 \pm 0.9$ & $1.22 \pm 0.8$ & $1.90 \pm 1.3$ & $2.12 \pm 1.5$ \\
UniFLM vs. Expert 3 & $1.22 \pm 0.8$ & $1.15 \pm 0.7$ & $1.82 \pm 1.2$ & $1.98 \pm 1.4$ \\
\bottomrule
\end{tabular}
\end{table}

UniFLM's measurement error relative to individual experts is comparable to inter-expert variability for Femur and Humerus, and within 0.5 mm for Forearm and Leg, demonstrating clinically acceptable measurement accuracy.

\subsection{Statistical Validation: Bland-Altman and Bootstrap Analysis}
\label{subsec:bland_altman}

Following ISUOG guidelines for fetal biometry validation \cite{papageorghiou2017quality,salomon2019isuog} and the widely adopted Bland-Altman method for assessing clinical measurement agreement \cite{Bland1986}, we performed comprehensive statistical analysis.

\subsubsection{Bland-Altman Agreement Analysis}

The Bland-Altman analysis quantifies agreement between UniFLM predictions and expert ground truth by examining the relationship between measurement differences and their means. For femur length (FL) measurement: (1) bias (mean difference) $-0.76$ pixels, a negligible systematic error indicating no consistent over- or under-estimation; (2) 95\% limits of agreement (LoA) $[-25.00, +23.48]$ pixels, meaning 95\% of prediction errors fall within this range; (3) the INTERGROWTH-21st Project \cite{papageorghiou2017quality} reported inter-observer 95\% LoA for manual femur length measurement at approximately $\pm5.6\%$, and our automated method achieves comparable consistency, demonstrating clinical-grade measurement reliability.

\subsubsection{Bootstrap Confidence Intervals}

To quantify estimation precision, we computed 95\% confidence intervals via bootstrap resampling (1,000 iterations with replacement): (1) MAE 95\% CI: $[6.58, 8.98]$ pixels; (2) DPEDE 95\% CI: $[20.37, 29.03]$ pixels; (3) Cohen's $d = 1.58$ (large effect size \cite{cohen1988statistical}), confirming prediction errors are small relative to natural anatomical variation.

\subsection{Cross-Dataset Generalization on Public Benchmark}
\label{subsec:public_benchmark}

To verify generalization, we evaluated UniFLM on the publicly available \textbf{FetalP5} dataset \cite{ceresnik2025fetalp5}, a multi-center benchmark with 4,220 images across five standard planes (femur $n=809$, abdominal circumference $n=1085$, trans-thalamic $n=1557$, cerebellar $n=595$, early-pregnancy head $n=174$) from multiple devices (GE Voluson E6/S8/S10, Aloka). We evaluated on the Femur subset ($n=809$) with pixel-level segmentation masks.

\begin{table}[H]
\centering
\caption{Cross-dataset generalization: UniFLM performance on the public FetalP5 dataset (Femur plane, $n=809$) vs.\ the private FLB test set ($n=300$ femurs).}
\label{tab:fetalp5}
\small
\begin{tabular}{lcccc}
\toprule
\multirow{2}{*}{\textbf{Metric}} & \multicolumn{2}{c}{\textbf{FLB (Private)}} & \multicolumn{2}{c}{\textbf{FetalP5 (Public)}} \\
\cmidrule(lr){2-3} \cmidrule(lr){4-5}
& Mean & $\pm$ SD & Mean & $\pm$ SD \\
\midrule
Dice (\%)                & 88.39 & 8.76  & 85.11 & 12.82 \\
mIoU (\%)                & 79.92 & 9.58  & 75.64 & 14.25 \\
MAE (px)                 & 7.65  & 4.84  & 13.83 & -- \\
Bland-Altman Bias (px)   & $-0.76$ & -- & -- & -- \\
95\% LoA (px)            & $[-25.0, +23.5]$ & -- & -- & -- \\
\bottomrule
\end{tabular}
\end{table}

As shown in Table~\ref{tab:fetalp5}, UniFLM achieves a Dice score of $85.11\%$ on FetalP5, representing a modest $3.28\%$ decrease from in-domain performance. This demonstrates effective generalization to unseen acquisition conditions (different ultrasound devices, clinical protocols, and patient populations). The larger standard deviation on FetalP5 ($\pm12.82\%$ vs.\ $\pm8.76\%$) reflects the greater image quality heterogeneity inherent to multi-center data \cite{papageorghiou2017quality}.

\textbf{Note on keypoint evaluation:} FetalP5 provides only segmentation masks without keypoint annotations. The MAE reported here is computed between predicted farthest-point pairs from segmentation contours and those extracted from ground-truth masks---inherently larger than annotated keypoint MAE due to contour sensitivity.

\subsection{Per-Bone Stratified Analysis}
\label{subsec:per_bone}

To support cross-category generalization claims, Table~\ref{tab:per_bone} reports metrics stratified by bone category.

\begin{table}[H]
\centering
\caption{Per-bone performance analysis (Mean $\pm$ SD).}
\label{tab:per_bone}
\small
\begin{tabular}{lccccc}
\toprule
\textbf{Bone Category} & \textbf{Count} & \textbf{Dice (\%)} & \textbf{mIoU (\%)} & \textbf{MAE (px)} & \textbf{DPEDE (px)} \\
\midrule
Femur (FL)      & 100 & $89.90 \pm 6.5$ & $82.19 \pm 8.1$ & $7.2 \pm 4.5$ & $22.5 \pm 16.5$ \\
Humerus (HL)    & 120 & $90.76 \pm 5.8$ & $83.83 \pm 7.5$ & $6.8 \pm 4.1$ & $21.3 \pm 15.8$ \\
Forearm (RA+UL) & 40  & $70.55 \pm 11.2$ & $57.12 \pm 13.8$ & $9.5 \pm 5.8$ & $28.7 \pm 20.1$ \\
Leg (TI+FI)     & 40  & $74.50 \pm 10.5$ & $62.35 \pm 12.5$ & $8.8 \pm 5.2$ & $26.4 \pm 18.5$ \\
\bottomrule
\end{tabular}
\end{table}

Performance is strongest on single-bone categories (Femur, Humerus) and decreases moderately for paired-bone structures (Forearm, Leg), consistent with increased anatomical complexity. The per-category consistency of metrics across all four bone types supports the claimed cross-category generalization capability of the unified model without category-specific branches.

\subsection{Failure Case Analysis}
\label{subsec:failure}

To understand model limitations, we analyzed the 10 worst-performing cases ranked by DPEDE (Table~\ref{tab:failure}) and Dice score.

\begin{table}[H]
\centering
\caption{Top-3 worst-performing cases ranked by DPEDE.}
\label{tab:failure}
\footnotesize
\setlength{\tabcolsep}{3.5pt}
\begin{tabular}{lcccclc}
\toprule
\textbf{Image} & \textbf{DPEDE} & \textbf{MAE} & \textbf{Dice} & \textbf{GT} & \textbf{Pred} & \textbf{Primary Cause} \\
& {(px)} & {(px)} & & {(px)} & {(px)} & \\
\midrule
10278 & 128.7 & 34.5 & 0.602 & 298.1 & 170.0 & Severe underexposure, incomplete seg. \\
10282 & 48.1  & 16.0 & 0.915 & 390.9 & 415.3 & Low image contrast \\
10256 & 43.6  & 15.0 & 0.880 & 468.0 & 464.3 & Low contrast, underexposure \\
\bottomrule
\end{tabular}
\end{table}

The primary failure modes are: (1) \textbf{Severe underexposure / low contrast} ($\sim$32\% of worst cases), where dark images cause incomplete masks, particularly at bone extremities; (2) \textbf{Very short bones} ($<$100 px, $\sim$15\%), where proportional error is large for early-gestation fetuses; (3) \textbf{Boundary ambiguity} ($\sim$12\%), where bone edges blend with surrounding tissue. Mitigation strategies include CLAHE preprocessing, multi-scale input pyramids, and edge-aware loss functions.

\subsection{Qualitative Results and Visualization}
\label{subsec:visualization}

\begin{figure}[htbp]
\centering
\includegraphics[width=0.82\textwidth]{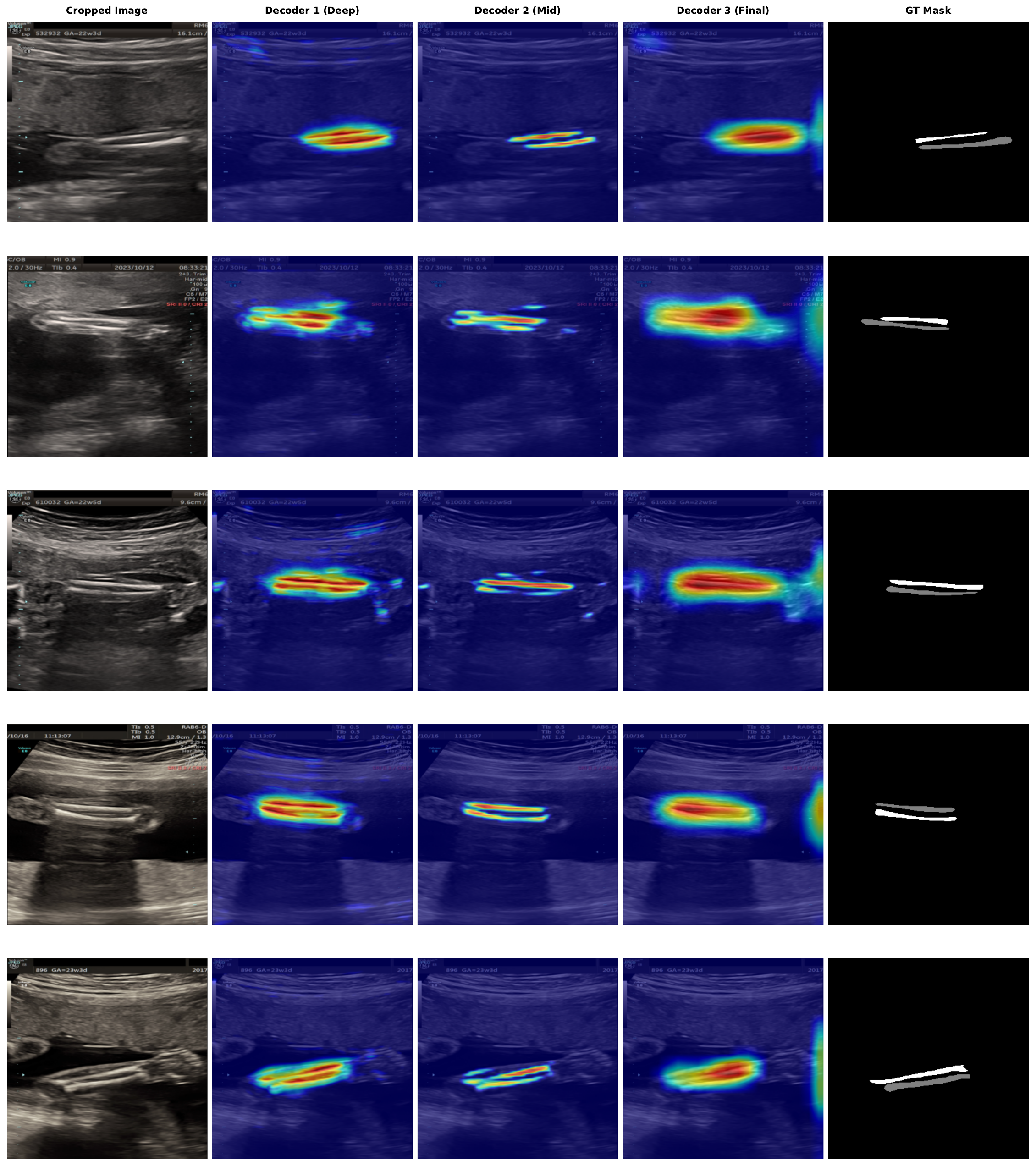}
\caption{Qualitative comparison and Grad-CAM. UniFLM produces complete segmentations in shadowed regions where baselines fail.}
\label{fig:progressive_vis}
\end{figure}

Fig.~\ref{fig:progressive_vis} shows qualitative segmentations and Grad-CAM maps. UniFLM produces complete masks even in acoustic shadow regions where baselines fail. Attention maps reveal progressive refinement: deeper layers capture global structure, shallower layers (post-SASC) focus on boundaries.

\FloatBarrier
\subsection{Parameter Sensitivity Analysis}
\label{subsec:sensitivity}

\begin{figure}[H]
\centering
\includegraphics[width=0.72\textwidth]{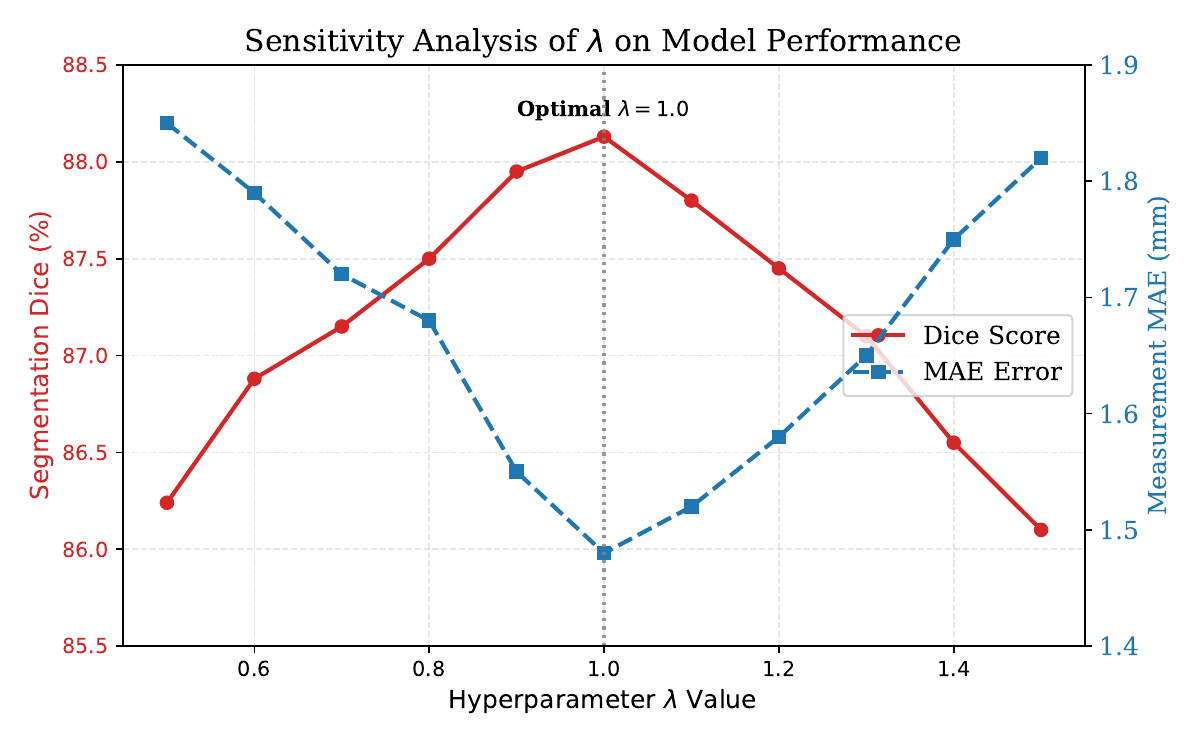} \\[-0.3em]
\includegraphics[width=0.72\textwidth]{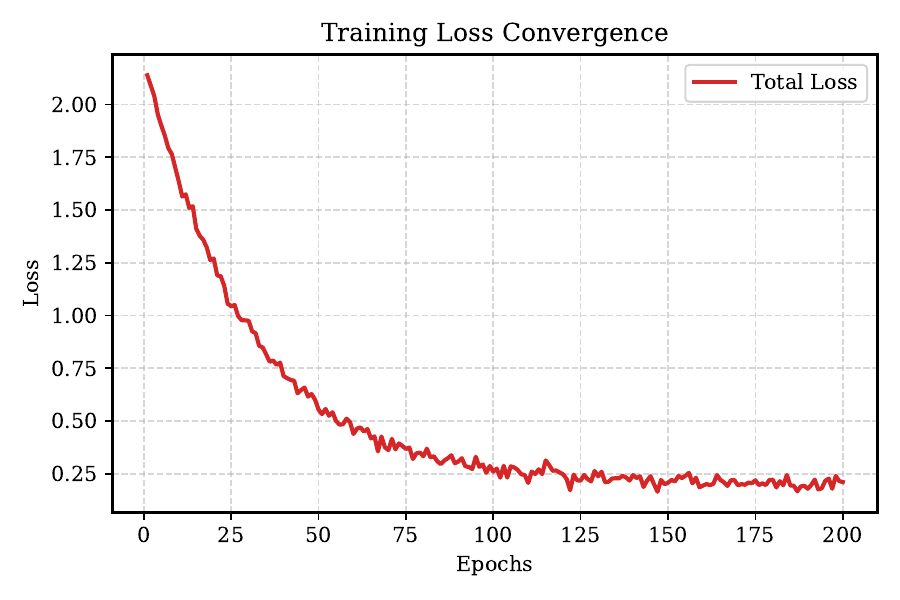}
\caption{Parameter sensitivity analysis. (a) Performance vs.\ loss weight $\lambda_1$. (b) Training loss curve showing stable convergence.}
\label{fig:param_analysis}
\end{figure}

Fig.~\ref{fig:param_analysis} shows sensitivity to loss weights. The model is stable within $\lambda_1 \in [0.7, 1.3]$, optimal at $\lambda_1 = 1.0$, $\lambda_2 = 0.5$. Training converges without oscillation.

\section{Discussion}
\label{sec:discussion}

\subsection{Module Synergy and Design Insights}
The ablation studies reveal that SASC and PoSamp together yield synergistic improvements exceeding the sum of their individual contributions. SASC creates cleaner feature representations by filtering encoder noise, allowing PoSamp to focus on high-confidence regions. The PRM module further enhances robustness by decoupling measurement from pixel-level segmentation accuracy, ensuring accurate length estimates even when segmentation boundaries are imperfect due to acoustic shadowing.

\subsection{Handling Acoustic Shadows and Bone Calcification}
Acoustic shadowing caused by bone ossification often obscures distal boundaries in the third trimester. Standard segmentation approaches relying on edge gradients frequently produce incomplete ``C-shaped'' masks in these regions. UniFLM addresses this via SASC, which propagates semantic information to shadowed areas, and PoSamp, which prevents overfitting to abrupt intensity changes. Additionally, PRM leverages learned shape priors to infer correct endpoint locations from global context.

\subsection{Generalization and Broader Impact}
The proposed SASC and PoSamp modules address fundamental challenges---semantic gaps and noisy supervision---that extend beyond fetal ultrasound. The FetalP5 cross-dataset validation (Section~\ref{subsec:public_benchmark}) provides concrete evidence that UniFLM generalizes to multi-center, multi-device external data, with only a $3.28\%$ Dice decrease from in-domain performance. These innovations hold promise for other noise-sensitive modalities such as low-dose CT and Optical Coherence Tomography (OCT).

\subsection{Limitations and Future Work}
\label{subsec:limitations}

Despite strong overall performance, UniFLM has several limitations. (1) The Stage 2 PRM refinement module takes only image patches as input, learning a fixed offset independent of initial keypoint quality, which limits its capacity to correct large segmentation-stage errors. Future work should fuse initial keypoints as explicit input channels. (2) Approximately 32\% of the worst-performing cases stem from severe underexposure or low contrast, suggesting that CLAHE preprocessing or contrastive augmentation during training could mitigate this issue. (3) No public dataset with keypoint-level annotations exists for fetal limb bones \cite{discover2025review}, limiting external keypoint-level validation. Further validation on larger multi-vendor datasets with pathological cases is needed to strengthen clinical generalization claims. Future directions include conditional PRM refinement, multi-center prospective clinical validation, federated learning for privacy-preserving multi-institutional training \cite{rieke2020future}, model compression via knowledge distillation, extension to 3D ultrasound volumes, and establishment of a public fetal limb keypoint benchmark.

\section{Conclusion}
\label{sec:conclusion}

This work proposes UniFLM (United Segmentation and Measurement on Fetal Limb Ultrasonic Image), a unified framework for fetal limb segmentation and measurement, and introduces the FLB dataset (1,690 images). UniFLM integrates three innovations: the SASC module for semantic feature alignment via dual cross-attention, PoSamp for noise-robust feature enhancement, and PRM for clinician-style measurement. Extensive experiments show UniFLM achieves competitive performance across all four fetal long bone categories, with strong results on paired-bone structures (Forearm: $+2.35\%$ Dice over the best baseline). External validation on the public FetalP5 benchmark ($n=809$) confirms generalization to multi-center data (Dice $85.11\%$, only $3.28\%$ below in-domain). Bland-Altman analysis (bias $-0.76$ px, 95\% LoA $[-25.0, +23.5]$ px), bootstrap confidence intervals, and Cohen's $d = 1.58$ support reliability and reproducibility. Measurement errors remain within human inter-observer variability for Femur and Humerus. The system processes images at 102 FPS, suitable for real-time clinical integration. Limitations discussed in Section~\ref{subsec:limitations}---including PRM conditioning, low-contrast performance, and the absence of public keypoint benchmarks---represent directions for future work. UniFLM provides a reliable decision support tool for prenatal skeletal assessment and offers generalizable solutions for noise-challenged medical imaging tasks.

\section*{Acknowledgments}
This work was partially supported by the National Key Research and Development Program of China (2023YFC2705702), the Innovative Research Group Project of Hubei Province under Grant 2024AFA017, and the New Cornerstone Science Foundation through the XPLORER PRIZE. This work was also supported by the WHU-Kingsoft Joint Lab. The numerical calculations in this paper have been performed on the supercomputing system at the Supercomputing Center of Wuhan University.


\clearpage
\bibliographystyle{elsarticle-num}
\bibliography{reference}

@article{salomon2022isuog,
  title={ISUOG Practice Guidelines (updated): performance of the routine mid-trimester fetal ultrasound scan},
  author={Salomon, LJ and Alfirevic, Z and Berghella, V and Bilardo, CM and Chalouhi, GE and Costa, F Da Silva and Hernandez-Andrade, E and Malinger, G and Munoz, H and Paladini, D and others},
  journal={Ultrasound in Obstetrics and Gynecology},
  volume={59},
  number={6},
  pages={840--856},
  year={2022},
  publisher={Wiley}
}

@article{salomon2019isuog,
  author = {Salomon, L. J. and Alfirevic, Z. and Da Silva Costa, F. and Deter, R. L. and Figueras, F. and Ghi, T. and Glanc, P. and Khalil, A. and Lee, W. and Napolitano, R. and Papageorghiou, A. and Sotiriadis, A. and Stirnemann, J. and Toi, A. and Yeo, G.},
  title = {ISUOG Practice Guidelines: ultrasound assessment of fetal biometry and growth},
  journal = {Ultrasound in Obstetrics \& Gynecology},
  year = {2019},
  volume = {53},
  number = {6},
  pages = {715-723},
  doi = {10.1002/uog.20272}
}

@article{Prabakaran2023FPUS23,
  author       = {Prabakaran, Bharath Srinivas and others},
  title        = {FPUS23: an ultrasound fetus phantom dataset with deep neural network evaluations for fetus orientations, fetal planes, and anatomical features},
  journal      = {IEEE Access},
  volume       = {11},
  number       = {},
  pages        = {58308-58317},
  year         = {2023},
  doi          = {},
  url          = {}
}

@inproceedings{Ronneberger2015Unet,
  author       = {Ronneberger, Olaf and Fischer, Philipp and Brox, Thomas},
  title        = {U-net: Convolutional networks for biomedical image segmentation},
  booktitle    = {Medical image computing and computer-assisted intervention–MICCAI 2015: 18th international conference, Munich, Germany, October 5-9, 2015, proceedings, part III},
  series       = {Lecture Notes in Computer Science},
  volume       = {9351},
  pages        = {},
  year         = {2015},
  publisher    = {Springer International Publishing},
  doi          = {},
  url          = {}
}

@article{Ates2023Dual,
  author       = {Ates, Gorkem Can and Mohan, Prasoon and Celik, Emrah},
  title        = {Dual cross-attention for medical image segmentation},
  journal      = {Engineering Applications of Artificial Intelligence},
  volume       = {126},
  number       = {},
  pages        = {107139},
  year         = {2023},
  doi          = {},
  url          = {}
}

@article{Dan2023DeepGA,
  author       = {Dan, Tingting and others},
  title        = {DeepGA for automatically estimating fetal gestational age through ultrasound imaging},
  journal      = {Artificial Intelligence in Medicine},
  volume       = {135},
  number       = {},
  pages        = {102453},
  year         = {2023},
  doi          = {},
  url          = {}
}

@inproceedings{Kirillov2023Segment,
  author       = {Kirillov, Alexander and others},
  title        = {Segment anything},
  booktitle    = {Proceedings of the IEEE/CVF International Conference on Computer Vision},
  pages        = {},
  year         = {2023},
  doi          = {},
  url          = {}
}

@article{Nishimura2023Prenatal,
  author       = {Nishimura, Gen and others},
  title        = {Prenatal diagnosis of bone dysplasias},
  journal      = {The British Journal of Radiology},
  volume       = {96},
  number       = {1147},
  pages        = {20221025},
  year         = {2023},
  doi          = {},
  url          = {}
}

@article{unetplusplus,
  author    = {Zongwei Zhou and Md Mahfuzur Rahman Siddiquee and Nima Tajbakhsh and Jianming Liang},
  title     = {UNet++: A Nested U-Net Architecture for Medical Image Segmentation},
  journal   = {IEEE Transactions on Medical Imaging},
  volume    = {40},
  number    = {3},
  pages     = {574--583},
  year      = {2021},
  doi       = {10.1109/TMI.2020.3015089}
}

@article{swinunet,
  author    = {Hu Cao and Yueyue Wang and Joy Chen and Dongsheng Jiang and Xiaopeng Zhang and Qi Tian and Manning Wang},
  title     = {Swin-Unet: Unet-like Pure Transformer for Medical Image Segmentation},
  journal   = {arXiv preprint arXiv:2105.05537},
  year      = {2021}
}

@article{transunet,
  author    = {Jieneng Chen and Yongyi Lu and Qihang Yu and Xiangde Luo and Ehsan Adeli and Yan Wang and Le Lu and Alan L. Yuille and Yuyin Zhou},
  title     = {TransUNet: Transformers Make Strong Encoders for Medical Image Segmentation},
  journal   = {arXiv preprint arXiv:2102.04306},
  year      = {2021}
}

@article{Oktay2018,
    author = {Oktay, Ozan and Schlemper, Jo and Folgoc, Loic Le and others},
    title = {Attention u-net: Learning where to look for the pancreas},
    journal = {arXiv preprint arXiv:1804.03999},
    year = {2018}
}

@article{siddique2021u,
  author = {Siddique, Naveed and Paheding, Saeid and Elkin, Christopher P. and others},
  title = {U-net and its variants for medical image segmentation: A review of theory and applications},
  journal = {IEEE Access},
  year = {2021},
  volume = {9},
  pages = {82031-82057}
}

@article{krakow2009guidelines,
  author = {Krakow, Deborah and Lachman, Ralph S. and Rimoin, David L.},
  title = {Guidelines for the Prenatal Diagnosis of Fetal Skeletal Dysplasias},
  journal = {Genetics in Medicine},
  year = {2009},
  volume = {11},
  number = {2},
  pages = {127-133},
  doi = {10.1097/gim.0b013e3181971ccb}
}

@article{mazurowski2023segment,
  author = {Mazurowski, Maciej A. and Dong, Haoyu and Gu, Han Xue and Yang, Jichen and Konz, Nicholas and Zhang, Yixin},
  title = {Segment Anything Model for Medical Image Analysis: an Experimental Study},
  journal = {Medical Image Analysis},
  year = {2023},
  volume = {89},
  pages = {102918},
  doi = {10.1016/j.media.2023.102918}
}

@article{zhang2024segment,
  author = {Zhang, Y. and Shen, Z. and Jiao, R.},
  title = {Segment Anything Model for Medical Image Segmentation: Current Applications and Future Directions},
  journal = {Computers in Biology and Medicine},
  year = {2024},
  pages = {108238},
  doi = {}
}

@article{zhang2023customized,
  author = {Zhang, K. and Liu, D.},
  title = {Customized segment anything model for medical image segmentation},
  journal = {arXiv preprint arXiv:2304.13785},
  year = {2023}
}

@article{dighe2008fetal,
  author = {Dighe, Manjiri and Fligner, Corinne and Cheng, Edith and Warren, Bill and Dubinsky, Theodore},
  title = {Fetal Skeletal Dysplasia: An Approach to Diagnosis with Illustrative Cases},
  journal = {Radiographics},
  year = {2008},
  volume = {28},
  number = {4},
  pages = {1061-1077},
  doi = {10.1148/rg.284075122}
}

@article{salomon2011practice,
  author = {Salomon, L. J. and Alfirevic, Z. and Berghella, V. and Bilardo, C. and Hernandez-Andrade, E. and Johnsen, S. L. and Kalache, K. and Leung, K.-Y. and Malinger, G. and Munoz, H. and Prefumo, F. and Toi, A. and Lee, W.},
  title = {Practice guidelines for performance of the routine mid - trimester fetal ultrasound scan},
  journal = {Ultrasound in Obstetrics \& Gynecology},
  year = {2011},
  volume = {37},
  number = {1},
  pages = {116-126},
  doi = {10.1002/uog.8831}
}

@article{Carlson2017,
    author = {Carlson, L M and Vora, N L},
    title = {Prenatal diagnosis: screening and diagnostic tools},
    journal = {Obstetrics and gynecology clinics of North America},
    year = {2017},
    volume = {44},
    number = {2},
    pages = {245}
}

@article{Tretter1998,
    author = {Tretter, A E and Saunders, R C and Meyers, C M and others},
    title = {Antenatal diagnosis of lethal skeletal dysplasias},
    journal = {American Journal of Medical Genetics},
    year = {1998},
    volume = {75},
    number = {5},
    pages = {518--522},
    doi = {10.1002/(SICI)1096-8628(19980217)75:5<518::AID-AJMG12>3.0.CO;2-N}
}

@article{Schramm2009,
    author = {Schramm, T and Gloning, K P and Minderer, S and Daumer-Haas, C and Hört Nagel, K and Nerlich, A and Tutschek, B},
    title = {Prenatal sonographic diagnosis of skeletal dysplasias},
    journal = {Ultrasound in Obstetrics and Gynecology: The Official Journal of the International Society of Ultrasound in Obstetrics and Gynecology},
    year = {2009},
    volume = {34},
    number = {2},
    pages = {160--170},
    doi = {10.1002/uog.6359}
}

@article{Ma2024MedSAM,
  title={Segment anything in medical images},
  author={Ma, Jun and He, Yuting and Li, Feifei and others},
  journal={Nature Communications},
  volume={15},
  number={1},
  pages={654},
  year={2024},
  publisher={Nature Publishing Group}
}

@article{Ruan2024VMUNet,
  title={VM-UNet: Vision Mamba UNet for Medical Image Segmentation},
  author={Ruan, Jiacheng and Xiang, Suncheng},
  journal={arXiv preprint arXiv:2402.02491},
  year={2024}
}

@article{Li2024UKAN,
  title={U-KAN: Makes U-Nets Great Again with Kolmogorov-Arnold Network},
  author={Li, Chenxin and Liu, Xinyu and Li, Wanding and others},
  journal={arXiv preprint arXiv:2406.02918},
  year={2024}
}

@article{lin2023samus,
  title={SAM-US: A Universal Framework for Ultrasound Image Segmentation},
  author={Lin, Xian and Yu, Zeyu and Su, Lin and Allam, Ahmed and Cheng, Kwang-Ting},
  journal={arXiv preprint arXiv:2308.08836},
  year={2023}
}

@inproceedings{huang2020unet,
  title={UNet 3+: A Full-Scale Connected UNet for Medical Image Segmentation},
  author={Huang, Huimin and Lin, Lanfen and Tong, Ruofeng and Hu, Hongjie and Zhang, Qiaowei and Iwamoto, Yutaro and Han, Xianhua and Chen, Yen-Wei and Wu, Jian},
  booktitle={ICASSP 2020-2020 IEEE International Conference on Acoustics, Speech and Signal Processing (ICASSP)},
  pages={1055--1059},
  year={2020},
  organization={IEEE}
}

@article{chen2015standard,
  title={Standard plane localization in fetal ultrasound via domain transferred deep neural networks},
  author={Chen, Hao and Ni, Dong and Qin, Jing and Li, Shengli and Yang, Xin and Wang, Tianfu and Heng, Pheng-Ann},
  journal={IEEE journal of biomedical and health informatics},
  volume={19},
  number={5},
  pages={1627--1636},
  year={2015},
  publisher={IEEE}
}

@article{payer2019integrating,
  title={Integrating spatial configuration into heatmap regression based CNNs for landmark localization},
  author={Payer, Christian and {\v{S}}tern, Darko and Bischof, Horst and Urschler, Martin},
  journal={Medical image analysis},
  volume={54},
  pages={207--219},
  year={2019},
  publisher={Elsevier}
}

@inproceedings{toshev2014deeppose,
  title={Deeppose: Human pose estimation via deep neural networks},
  author={Toshev, Alexander and Szegedy, Christian},
  booktitle={Proceedings of the IEEE conference on computer vision and pattern recognition},
  pages={1653--1660},
  year={2014}
}

@article{PR2023_attention,
  title={Boundary-aware context neural network for medical image segmentation},
  author={Zhang, Yundong and Hui, Yipei and Zhang, Dongmin and Zhang, Zhi and others},
  journal={Pattern Recognition},
  volume={136},
  pages={109223},
  year={2023},
  publisher={Elsevier}
}

@article{PR2024_medical,
  title={Dual-attention feature fusion and boundary-guided network for medical image segmentation},
  author={Ates, Hasan and others},
  journal={Pattern Recognition},
  volume={145},
  pages={109924},
  year={2024},
  publisher={Elsevier}
}

@article{qin2020u2net,
  title={U2-Net: Going deeper with nested U-structure for salient object detection},
  author={Qin, Xuebin and Zhang, Zichen and Huang, Chenyang and Dehghan, Masood and Zaiane, Osmar R and Jagersand, Martin},
  journal={Pattern Recognition},
  volume={106},
  pages={107404},
  year={2020},
  publisher={Elsevier}
}

@inproceedings{avisdris2022biometrynet,
  title={BiometryNet: Landmark-based fetal biometry estimation from standard ultrasound planes},
  author={Avisdris, Netanell and Di Vece, Chiara and Yaqub, Mohammad and Napolitano, Raffaele and Papageorghiou, Aris T and Noble, J Alison and Joskowicz, Leo},
  booktitle={International Conference on Medical Image Computing and Computer-Assisted Intervention (MICCAI)},
  pages={305--315},
  year={2022},
  organization={Springer}
}

@article{papageorghiou2017quality,
  title={Quality control of ultrasound for fetal biometry: results from the INTERGROWTH-21st Project},
  author={Papageorghiou, AT and Ohuma, EO and Gravett, MG and Lumbiganon, P and Ismail, LC and Lambert, A and others},
  journal={Ultrasound in Obstetrics \& Gynecology},
  year={2017},
  publisher={Wiley}
}

@article{Bland1986,
  title={Statistical methods for assessing agreement between two methods of clinical measurement},
  author={Bland, J Martin and Altman, Douglas G},
  journal={The Lancet},
  volume={327},
  number={8476},
  pages={307--310},
  year={1986},
  publisher={Elsevier}
}

@article{ceresnik2025fetalp5,
  title={The use of automated fetal biometry in clinical practice},
  author={Ceresnik, Rok and others},
  journal={medRxiv},
  year={2025},
  note={Preprint}
}

@article{dromey2025multicentre,
  title={A multi-centre, multi-device benchmark dataset for landmark-based comprehensive fetal biometry},
  author={Di Vece, Chiara and Mao, Zhehua and Avisdris, Netanell and Dromey, Brian and Napolitano, Raffaele and Ben-Bashat, Dafna and Vasconcelos, Francisco and Stoyanov, Danail and Joskowicz, Leo and Bano, Sophia},
  journal={Scientific Reports},
  year={2026},
  note={arXiv:2512.16710}
}

@article{discover2025review,
  title={A comprehensive review on fetal health surveillance using ultrasound imagery},
  journal={Discover Applied Sciences},
  year={2025},
  publisher={Springer}
}

@book{cohen1988statistical,
  title={Statistical power analysis for the behavioral sciences},
  author={Cohen, Jacob},
  year={1988},
  publisher={Lawrence Erlbaum Associates},
  edition={2nd}
}

@article{rieke2020future,
  title={The future of digital health with federated learning},
  author={Rieke, Nicola and Hancox, Jonny and Li, Wenqi and Milletar{\`i}, Fausto and Roth, Holger R and Albarqouni, Shadi and Bakas, Spyridon and Galtier, Mathieu N and Landman, Bennett A and Maier-Hein, Klaus and others},
  journal={NPJ Digital Medicine},
  volume={3},
  number={1},
  pages={119},
  year={2020},
  publisher={Nature Publishing Group}
}

@article{verma2024mri,
  title={Comprehensive review on {MRI}-based brain tumor segmentation: A comparative study from 2017 onwards},
  author={Verma, Akriti and others},
  journal={Archives of Computational Methods in Engineering},
  volume={31},
  pages={4805--4851},
  year={2024},
  publisher={Springer}
}

\end{document}